\documentclass[11pt,a4paper]{article}
\usepackage{times}
\usepackage{latexsym}

\usepackage{microtype}

\usepackage{booktabs}
\usepackage{caption}
\usepackage{color}
\usepackage{amsfonts}
\usepackage{amsmath}
\usepackage{algorithm}
\usepackage{algorithmic}
\usepackage{graphicx}
\usepackage{nicefrac}
\usepackage{bbm}
\usepackage{amsthm}
\usepackage{wrapfig}
\usepackage{tikz}
\usepackage[titletoc,title]{appendix}
\usepackage{stmaryrd}
\usepackage{amsmath}
\usepackage{amssymb}
\usepackage{url}
\usepackage[hyperref]{acl2020}

\aclfinalcopy 

\newcommand{\ourmodel}{SenseBERT}
\newcommand{\ourmodelbase}{SenseBERT$_{\textsc{base}}$}
\newcommand{\ourmodellarge}{SenseBERT$_{\textsc{large}}$}
\newcommand{\bertbase}{BERT$_{\textsc{base}}$}
\newcommand{\bertlarge}{BERT$_{\textsc{large}}$}

\newcommand{\R}{{\mathbb R}}
\newcommand{\eg}{\emph{e.g.}}
\newcommand{\ie}{\emph{i.e.}}

\newcommand{\etc}{\emph{etc.}}

\title{Sense{BERT}: Driving Some Sense into {BERT}}

\author{ Yoav Levine~~~~Barak Lenz~~~~Or Dagan~~~~Ori Ram~~~~Dan Padnos~~~~Or Sharir \\  \textbf{Shai Shalev-Shwartz~~~~Amnon Shashua~~~~Yoav Shoham} \\ \\
AI21 Labs, Tel Aviv, Israel \\ \\
\texttt{\{yoavl,barakl,ord,orir,...\}@ai21.com}
}

\begin{document}
	\maketitle
	
	\begin{abstract}
		The ability to learn from large unlabeled corpora has allowed neural language models to advance the frontier in
		natural language understanding. However, existing self-supervision techniques operate at the word {\it form} level, which serves as a surrogate for the underlying semantic content. This paper proposes a method to employ weak-supervision directly at the word \emph{sense} level. Our model, named SenseBERT, is pre-trained to predict not only the masked words but also their WordNet supersenses. Accordingly, we attain a lexical-semantic level language model, without the use of human annotation. SenseBERT achieves significantly improved lexical understanding, as we demonstrate by experimenting on SemEval Word Sense Disambiguation, and by attaining a state of the art result on the `Word in Context' task. 
	\end{abstract}

	\section{Introduction}\label{sec:intro}
	
	Neural language models have recently undergone a qualitative leap forward, pushing the state of the art on various NLP tasks. 
	Together with advances in network architecture \cite{vaswani2017attention}, the use of self-supervision has proven to be central to these achievements, as it allows the network to learn from massive amounts of unannotated text.

	The self-supervision strategy employed in BERT  \citep{devlin2018bert} involves masking some of the words in an input sentence, and then training the model to predict them given their context. 
	Other proposed approaches for self-supervised objectives, including unidirectional~\citep{radford2019language}, permutational~\citep{yang2019xlnet}, or word insertion-based~\citep{chan2019kermit} methods,  operate similarly, over words. 
	However, since a given word form can possess multiple meanings (\eg,~the word `bass' can refer to a fish, a guitar, a type of singer, \etc), the word itself is merely a surrogate of its actual meaning in a given context, referred to as its \emph{sense}.
	Indeed, the word-form level is viewed as a surface level which often introduces challenging ambiguity~\cite{navigli2009word}.
	
	In this paper, we bring forth a novel methodology for applying weak-supervision directly on the level of a word's meaning. 
	By infusing word-sense information into BERT's pre-training signal, we explicitely expose the model to lexical semantics when learning from a large unannotated corpus. We call the resultant sense-informed model \emph{{\ourmodel}}. 
	Specifically, we add a masked-word sense prediction task as an auxiliary task in BERT’s pre-training.
	Thereby, jointly with the standard word-form level language model, we train a \emph{semantic-level language model} that predicts the missing word’s meaning.
	Our method does not require sense-annotated data; self-supervised learning from unannotated text is facilitated by using WordNet~\citep{miller1998wordnet}, an expert constructed inventory of word senses, as weak supervision.

	We focus on a coarse-grained variant of a word's sense, referred to as its WordNet \emph{supersense}, in order to mitigate an identified brittleness of fine-grained word-sense systems, caused by arbitrary sense granularity, blurriness, and general subjectiveness ~\citep{kilgarriff1997don,schneider2014lexical}.  WordNet lexicographers organize all word senses into $45$ supersense categories, $26$ of which are for nouns, $15$ for verbs, $3$ for adjectives and $1$ for adverbs (see full supersense table in the supplementary materials).
	Disambiguating a word's supersense has been widely studied as a fundamental lexical categorization task ~\citep{ciaramita2003supersense,basile2012super,schneider2015corpus}.

	We employ the masked word's allowed supersenses list from WordNet as a set of possible labels for the sense prediction task. The labeling of words with a single supersense (\eg,~`sword' has only the supersense noun.artifact) is straightforward: We train the network to predict this supersense given the masked word's context. As for words with multiple supersenses (\eg,~`bass' can be: noun.food, noun.animal, noun.artifact, noun.person, \etc), we train the model to predict any of these senses, leading to a simple yet effective soft-labeling scheme. 
	
	We show that
	SenseBERT$_{\textsc{base}}$~outscores both \bertbase~and \bertlarge~by a large margin on a supersense variant of the SemEval Word Sense Disambiguation (WSD) data set standardized in~\citet{raganato2017word}. 
	Notably, \ourmodel~receives competitive results on this task without fune-tuning, \ie, when training a linear classifier over the pretrained embeddings, which serves as a testament for its self-acquisition of lexical semantics.
	Furthermore, we show that \ourmodelbase~surpasses \bertlarge~in the Word in Context (WiC) task~\citep{pilehvar2018wic} from  the SuperGLUE benchmark~\citep{wang2019superglue}, which directly depends on word-supersense awareness. 
	A single \ourmodellarge~model achieves state of the art performance on WiC with a score of $72.14$, improving the score of \bertlarge~by $2.5$ points.  
	
	\begin{figure*}
		\centering
		\includegraphics[width=\linewidth]{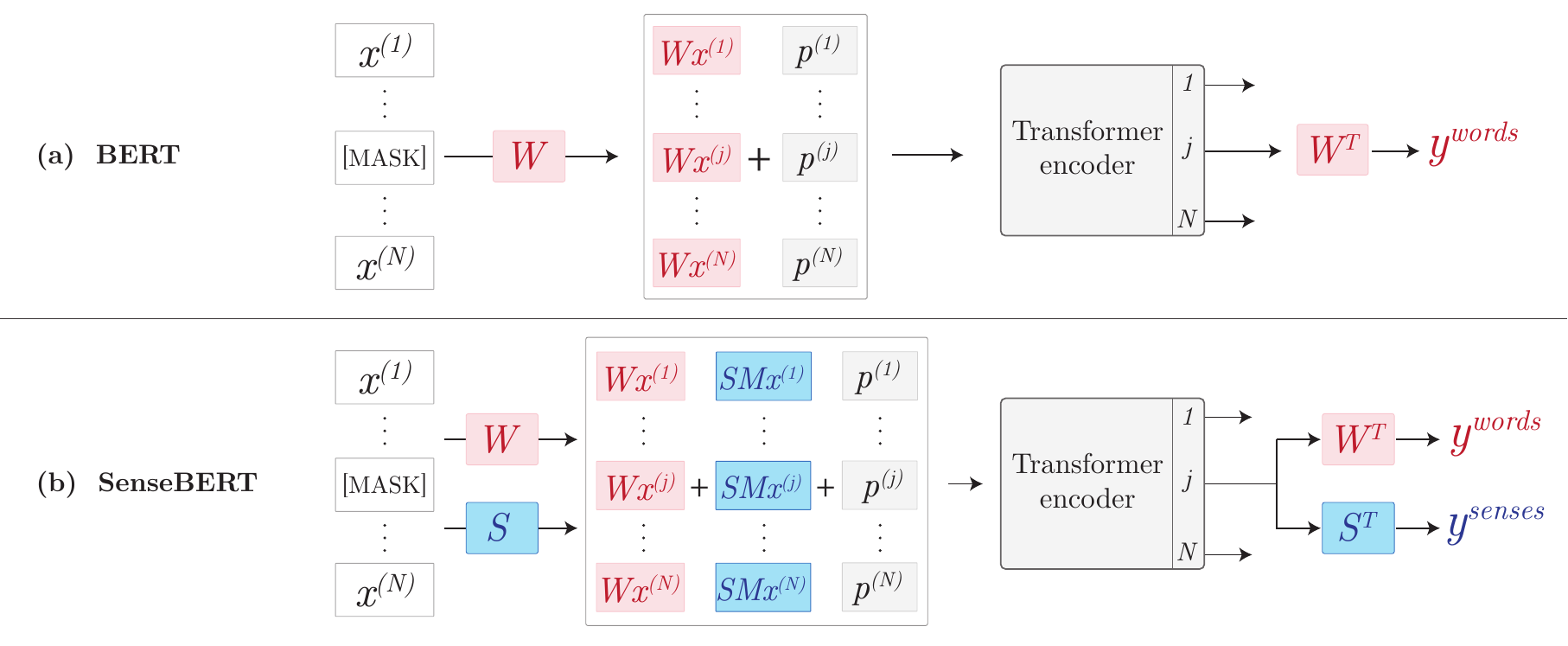}
		\vspace{-2mm} 
		\caption{\ourmodel~includes a masked-word supersense prediction task, pre-trained jointly with BERT's original masked-word prediction task~\citep{devlin2018bert} (see section~\ref{sec:method:loss}). As in the original BERT, the mapping from the Transformer dimension to the external dimension is the same  both at input and at output ($W$ for words and $S$ for supersenses), where $M$ denotes a fixed mapping between word-forms and their allowed WordNet supersenses (see section~\ref{sec:method:input}).
			The vectors $p^{(j)}$ denote positional embeddings. 
			For clarity, we omit a reference to a sentence-level Next Sentence Prediction task trained jointly with the above.}
		\label{fig:UVfig}
	\end{figure*}
	
%
\section{Related Work}\label{sec:related_work}

Neural network based word embeddings first appeared as a static mapping (non-contextualized), where every word is represented by a constant pretrained embedding~\citep{mikolov2013distributed,pennington2014glove}. 
Such embeddings were shown to contain some amount of word-sense information~\citep{iacobacci2016embeddings,yuan2016semi,arora2018linear,le2018deep}.
Additionally, sense embeddings computed for each word sense in the word-sense inventory
(e.g. WordNet) have been employed, relying on hypernymity relations~\citep{rothe2015autoextend} or the gloss for each sense~\citep{chen2014unified}. These approaches rely on static word embeddings and require a large amount of annotated data per word sense.

The introduction of contextualized word embeddings~\citep{peters2018deep}, for which a given word's embedding is context-dependent rather than precomputed, has brought forth a promising prospect for sense-aware word embeddings. Indeed, visualizations in~\citet{coenen2019visualizing} show that sense sensitive clusters form in BERT's word embedding space.
Nevertheless, we identify a clear gap in this abilty. We show that a vanilla BERT model trained with the current word-level self-supervision, burdened with the implicit task of disambiguating word meanings, often fails to grasp lexical semantics, exhibiting high supersense misclassification rates. Our suggested weakly-supervised word-sense signal allows \ourmodel~to significantly bridge this gap.

Moreover, \ourmodel~exhibits an improvement in lexical semantics ability (reflected by the Word in Context task score) even when compared to models with WordNet infused linguistic knowledge. Specifically we compare to \citet{peters2019knowledge} who re-contextualize word embeddings via a word-to-entity attention mechanism (where entities are WordNet lemmas and synsets), and to~\citet{loureiro2019language} which construct sense embeddings from BERT's word embeddings and use the WordNet graph to enhance coverage (see quantitative comparison in table~\ref{tab:wic}).

	\section{Incorporating Word-Supersense Information in Pre-training}
	\label{sec:method}
	In this section, we present our proposed method for integrating word sense-information within \ourmodel's pre-training.
	We start by describing the vanilla BERT architecture in subsection~\ref{sec:method:background}. We conceptually divide it into an internal transformer encoder and an external mapping  $W$ which translates the observed vocabulary space into and out of the transformer encoder space [see illustration in figure~\hyperref[fig:UVfig]{~\ref{fig:UVfig}(a)}]. 
	
		In the subsequent subsections, we frame our contribution to the vanilla BERT architecture as an addition of a parallel external mapping to the words supersenses space, denoted $S$ [see illustration in figure~\hyperref[fig:UVfig]{~\ref{fig:UVfig}(b)}]. Specifically, in section~\ref{sec:method:loss} we describe the loss function used for learning $S$ in parallel to $W$, effectively implementing word-form and word-sense multi-task learning in the pre-training stage. Then, in section~\ref{sec:method:input} we describe our methodology for adding supersense information in $S$ to the initial Transformer embedding, in parallel to word-level information added by $W$. 
		In section~\ref{sec:method:vocab} we address the issue of supersense prediction for out-of-vocabulary words, and in section~\ref{sec:method:single} we describe our modification of BERT's masking strategy, prioritizing single-supersensed words which carry a clearer semantic signal.
	
	\subsection{Background}\label{sec:method:background}
	The input to BERT is a sequence of words $\{x^{(j)} \in \{0,1\}^{D_W}\}_{j=1}^N$ where $15\%$ of the words are replaced by a [MASK] token (see treatment of sub-word tokanization in section~\ref{sec:method:vocab}).
	Here $N$ is the input sentence length, $D_W$ is  the word vocabulary size, and $x^{(j)}$ is a 1-hot vector corresponding to the $j^{\textrm {th}}$ input word.
	For every masked word, the output of the pretraining task is a word-score vector $y^{\textrm{words}}\in\R^{D_W}$ containing the per-word score. 
	BERT's architecture  can be decomposed to \textbf{(1)} an internal Transformer encoder architecture~\citep{vaswani2017attention} wrapped by \textbf{(2)} an external mapping to the word vocabulary space, denoted by $W$.\footnote{For clarity, we omit a description of the Next Sentence Prediction task which we employ as in \citet{devlin2018bert}.} 
	
	The Transformer encoder operates over a sequence of word embeddings $v_{\textrm {input}}^{(j)}\in\R^d$, where $d$ is the Transformer encoder's hidden dimension. These are passed through multiple attention-based Transformer layers, producing a new sequence
	of contextualized embeddings at each layer. The Transformer encoder output is the final sequence
	of contextualized word embeddings $v_{\textrm {output}}^{(j)}\in\R^d$. 
	
	The external mapping $W\in\R^{d\times D_W}$ is effectively a translation between 
	the external word vocabulary dimension and the internal Transformer dimension. 
	Original words in the input sentence are translated into the Transformer block by applying this mapping (and adding positional encoding vectors $p^{(j)}\in\R^d$):
	\begin{equation}\label{eq:input_old}
	v_{\textrm {input}}^{(j)}=Wx^{(j)} +p^{(j)}
	\end{equation}
	The word-score vector for a masked word at position $j$ is extracted from the Transformer encoder output by applying the transpose: $y^{\textrm{words}}=W^{\top}v_{\textrm {output}}^{(j)}$  [see illustration in figure~\hyperref[fig:UVfig]{~\ref{fig:UVfig}(a)}].
	The use of the same matrix $W$ as the mapping in and out of the transformer encoder space is referred to as  \emph{weight tying}~\citep{inan2016tying,press2016using}.
	
	Given a masked word in position $j$, BERT's original masked-word prediction pre-training task is to have the softmax of the word-score vector $y^{\textrm{words}}=W^{\top}v_{\textrm {output}}^{(j)}$ get as close as possible to a 1-hot vector corresponding to the masked word. 
	This is done by minimizing the cross-entropy loss between the softmax of the word-score vector and a 1-hot vector corresponding to the masked word:
	\begin{equation} \label{eq:reg_loss}
	\mathcal{L}_{\textrm{LM}}=-\log p(w|\textrm{context}),
	\end{equation}
	where $w$ is the masked word, the context is composed of the rest of the input sequence, and the probability is computed by:
	\begin{equation} 
	p(w|context)=\frac
	{\exp\left({y^{\textrm{words}}_w}\right)}
	{\sum_{w'}\exp\left({y^{\textrm{words}}_{w'}}\right)},
	\end{equation}
	where $y^{\textrm{words}}_w$ denotes the $w^{\textrm{th}}$ entry of the word-score vector.

	\subsection{Weakly-Supervised Supersense Prediction Task}
	\label{sec:method:loss}
	
	Jointly with the above procedure for training the word-level language model of \ourmodel, we train the model to predict the supersense of every masked word, thereby training a semantic-level language model. This is done by adding a parallel external mapping to the words supersenses space, denoted $S\in \R^{d\times D_S}$ [see illustration in figure~\hyperref[fig:UVfig]{~\ref{fig:UVfig}(b)}],  where $D_S=45$ is the size of supersenses vocabulary.
	Ideally, the objective is to have the softmax of the sense-score vector $y^{\textrm{senses}}\in\R^{D_S}:=S^{\top}v_{\textrm {output}}^{(j)}$ get as close as possible to a 1-hot vector corresponding to the word's supersense in the given context. 
	
	For each word $w$ in our vocabulary, we employ the WordNet word-sense inventory for constructing $A(w)$, the set of its ``allowed" supersenses. Specifically, we apply a WordNet Lemmatizer on $w$, extract the different synsets that are mapped to the lemmatized word in WordNet, and define $A(w)$ as the union of supersenses coupled to each of these synsets. As exceptions, we set $A(w)=\varnothing$ for the following: (i) short words (up to 3 characters), since they are often treated as abbreviations, (ii) stop words, as WordNet does not contain their main synset (e.g. `he' is either the element helium or the hebrew language according to WordNet), and (iii) tokens that represent part-of-word (see section~\ref{sec:method:vocab} for further discussion on these tokens).
	
    Given the above construction, we employ a combination of two loss terms for the supersense-level language model. The following \emph{allowed-senses term} maximizes the probability that the predicted sense is in the set of allowed supersenses of the masked word $w$:
	\begin{align} 
	\mathcal{L}_{\textrm{SLM}}^{\textrm{allowed}} &=-\log p\left(s\in A(w)|\textrm{context}\right) \nonumber\\
	&= -\log\sum_{s\in A(w)}p(s|\textrm{context}),
	\end{align}
	where the probability for a supersense $s$ is given by:	
	\begin{equation}
	p(s|\textrm{context}) = \frac
	{\exp(y^{\textrm{senses}}_s)}
	{\sum_{s'}\exp(y^{\textrm{senses}}_{s'})}.
	\end{equation}
	
	The soft-labeling scheme given above, which treats all the allowed supersenses of the masked word equally, introduces noise to the supersense labels. 
	We expect that encountering many contexts in a sufficiently large corpus will reinforce the correct labels whereas the signal of incorrect labels will diminish. 
	To illustrate this, consider the following examples for the food context:
	
	\begin{enumerate} 
		\item ``This \textbf{bass} is delicious" \\(supersenses: noun.food, noun.artifact, \etc) \item  ``This \textbf{chocolate} is delicious" \\(supersenses: noun.food, noun.attribute, \etc)
		\item  ``This \textbf{pickle} is delicious" \\(supersenses: noun.food, noun.state, \etc)
	\end{enumerate} 
	Masking the marked word in each of the examples results in three identical input sequences, each with a different sets of labels. The ground truth label, noun.food, appears in all cases, so that its probability in contexts indicating food is increased whereas the signals supporting other labels cancel out.

	While $\mathcal{L}_{\textrm{SLM}}^{\textrm{allowed}}$ pushes the network in the right direction, minimizing this loss could result in the network becoming overconfident in predicting a strict subset of the allowed senses for a given word, i.e., a collapse of the prediction distribution. This is especially acute in the early stages of the training procedure, when the network could converge to the noisy signal of the soft-labeling scheme.
	
	To mitigate this issue, the following \emph{regularization term} is added to the loss, which encourages a uniform prediction distribution over the allowed supersenses:
	\begin{equation} 
	\mathcal{L}_{\textrm{SLM}}^{\textrm{reg}}=-\sum_{s\in A(w)} \frac{1}{|A(w)|} \log p(s|\textrm{context}),
	\end{equation}
	\ie, a cross-entropy loss with a uniform distribution over the allowed supersenses.
	
	\begin{figure*}
		\centering
		\includegraphics[width=\linewidth]{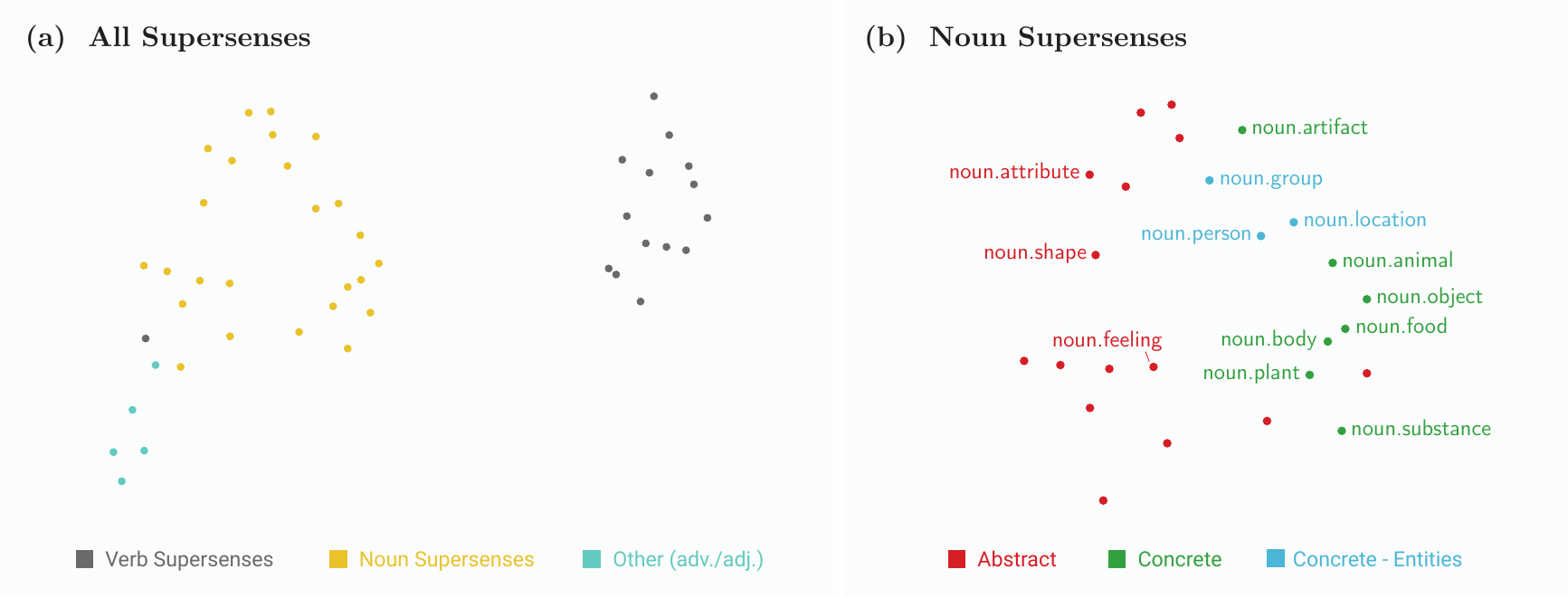}
		\vspace{-2mm} 
		\caption{UMAP visualization of supersense vectors (rows of the classifier $S$) learned by \ourmodel~at pre-training. \textbf{(a)} Clustering by the supersense's part-of speech. \textbf{(b)} Within noun supersenses, semantically similar supersenses are clustered together (see more details in the supplementary materials).}
		\label{fig:cluster}
	\end{figure*}
	
	Overall, jointly with the regular word level language model trained with the loss in eq.~\ref{eq:reg_loss}, we train the semantic level language model with a combined loss of the form:
	
	\begin{equation}\label{eq:slm_loss}
	\mathcal{L}_{\textrm{SLM}}=\mathcal{L}_{\textrm{SLM}}^{\textrm{allowed}}+\mathcal{L}_{\textrm{SLM}}^{\textrm{reg}}~~~~.	\end{equation}

	\subsection{Supersense Aware Input Embeddings}
	\label{sec:method:input}
	Though in principle two different matrices could have been used for converting in and out of the Tranformer encoder, the BERT architecture employs the same mapping $W$. 
	This approach, referred to as weight tying, was shown to yield theoretical and pracrical benefits~\cite{inan2016tying,press2016using}. 
	Intuitively, constructing the Transformer encoder's input embeddings from the same mapping with which the scores are computed improves their quality as it makes the input more sensitive to the training signal.
	
	We follow this approach, and insert our newly proposed semantic-level language model matrix $S$ in the input in addition to $W$ [as depicted in figure~\hyperref[fig:UVfig]{~\ref{fig:UVfig}(b)}], such that the input vector to the Transformer
	encoder (eq.~\ref{eq:input_old}) is modified to obey: 
	\begin{equation}\label{eq:input}
	v_{\textrm {input}}^{(j)}=(W+SM)x^{(j)} +p^{(j)},
	\end{equation}
	where $p^{(j)}$ are the regular positional embeddings as used in BERT, and $M\in\R^{D_S\times D_W}$ is a static 0/1 matrix converting between words and their allowed WordNet supersenses $A(w)$ (see construction details above).
	
	The above strategy for constructing $v_{\textrm {input}}^{(j)}$ allows for the semantic level vectors in $S$ to come into play and shape the input embeddings even for words which are rarely observed in the training corpus. For such a word, the corresponding row in $W$ is potentially less informative, since due to the low word frequency the model did not have sufficient chance to adequately learn it. 
	However, since the model learns a representation of its supersense, the  corresponding row in $S$ is informative of the semantic category of the word. Therefore, the input embedding in eq.~\ref{eq:input} can potentially help the model to elicit meaningful information even when the masked word is rare, allowing for better exploitation of the training corpus.
	
	\begin{figure*}
		\centering
		\includegraphics[width=\linewidth]{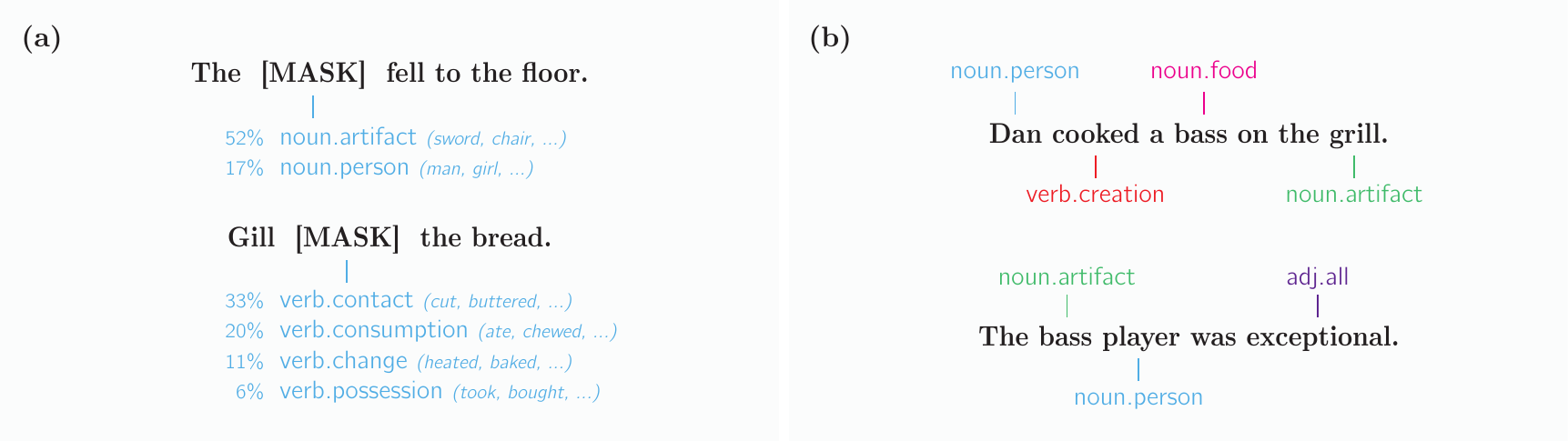}
		\vspace{-2mm} 
		\caption{\textbf{(a)} A demonstration of supersense probabilities assigned to a masked position within context, as given by \ourmodel's word-supersense level semantic language model (capped at $5\%$). Example words corresponding to each supersense are presented in parentheses. \textbf{(b)} Examples of \ourmodel's prediction on raw text, when the unmasked input sentence is given to the model. This beyond word-form abstraction ability facilitates a more natural elicitation of semantic content at pre-training.}
		\label{fig:slm}
	\end{figure*}
	
	\subsection{Rare Words Supersense Prediction}\label{sec:method:vocab}
	At the pre-processing stage, when an out-of-vocabulary (OOV) word is encountered in the corpus, it is divided into several in-vocabulary sub-word tokens. 
	For the self-supervised word prediction task (eq.~\ref{eq:reg_loss}) masked sub-word tokens are straightforwardly predicted as described in section~\ref{sec:method:background}. In contrast, word-sense supervision is only meaningful at the word level. We compare two alternatives for dealing with tokenized OOV words for the supersense prediction task (eq.~\ref{eq:slm_loss}).
	
	In the first alternative, called \emph{$60$K vocabulary}, we augment BERT's original $30$K-token vocabulary (which roughly contained the most frequent words) with additional 30K new words, chosen according to their frequency in Wikipedia.
	This vocabulary increase allows us to see more of the corpus as whole words for which supersense prediction is a meaningful operation. 
	Additionally, in accordance with the discussion in the previous subsection, our sense-aware input embedding mechanism can help the model extract more information from lower-frequency words. 
	For the cases where a sub-word token is chosen for masking, we only propagate the regular word level loss and do not train the supersense prediction task.

	The above addition to the vocabulary results in an increase of approximately $23$M parameters over the $110$M parameters of \bertbase~and an increase of approximately $30$M parameters over the $340$M parameters of \bertlarge ~(due to different embedding dimensions $d=768$ and $d=1024$, respectively). It is worth noting that similar vocabulary sizes in leading models have not resulted in increased sense awareness, as reflected for example in the WiC task results~\citep{liu2019roberta}.
	
	As a second alternative, referred to as \emph{average embedding}, we employ BERT's regular $30$K-token vocabulary and employ a whole-word-masking strategy. Accordingly, all of the tokens of a tokenized OOV word are masked together. In this case, we train the supersense prediction task to predict the WordNet supersenses of this word from the \emph{average} of the output embeddings at the location of the masked sub-words tokens. 
	

	\subsection{Single-Supersensed Word Masking}\label{sec:method:single}
	
	Words that have a single supersense are good anchors for obtaining an unambiguous semantic signal. These words teach the model to accurately map contexts to supersenses, such that it is then able to make correct context-based predictions even when a masked word has several supersenses. We therefore favor such words in the masking strategy, choosing $50\%$ of the single-supersensed words in each input sequence to be masked. We stop if $40\%$ of the overall $15\%$ masking budget is filled with single-supersensed words (this rarly happens), and in any case we randomize the choice of the remaining words to complete this budget. 
	As in the original BERT, $1$ out of $10$ words chosen for masking is shown to the model as itself rather than replaced with [MASK].
	
			\begin{table}[t]
		\begin{center}
			\begin{tabular}{lc}
				\toprule
				\ourmodelbase & \bf SemEval-SS Fine-tuned \\
				\midrule
				$30$K  no OOV & 81.9  \\
				$30$K  average OOV  & 82.7\\
				$60$K  no OOV & 83 \\
				\bottomrule
			\end{tabular}
		\end{center}
		\caption{Testing variants for predicting supersenses of rare words during~\ourmodel's pretraining, as described in section~\ref{sec:results:oov}. Results are reported on the SemEval-SS task (see section~\ref{sec:results:sem}). $30$K/$60$K stand for vocabulary size, and no/average OOV stand for not predicting senses for OOV words or predicting senses from the average of the sub-word token embeddings, respectively.}
		\label{tab:oov}
	\end{table} 
	
	\section{Semantic Language Model Visualization}
	\label{sec:visu}
	A \ourmodel~pretrained as described in section~\ref{sec:method} (with training hyperparameters as in~\citet{devlin2018bert}), has an immediate non-trivial bi-product. The pre-trained mapping to the supersenses space, denoted $S$, acts as an additional head predicting a word's supersense given context [see figure~\hyperref[fig:UVfig]{~\ref{fig:UVfig}(b)}]. 
	We thereby effectively attain a semantic-level language model that predicts the missing word's meaning
	jointly with the standard word-form level language model.
	
	\begin{figure*}
		\centering
		\includegraphics[width=\linewidth]{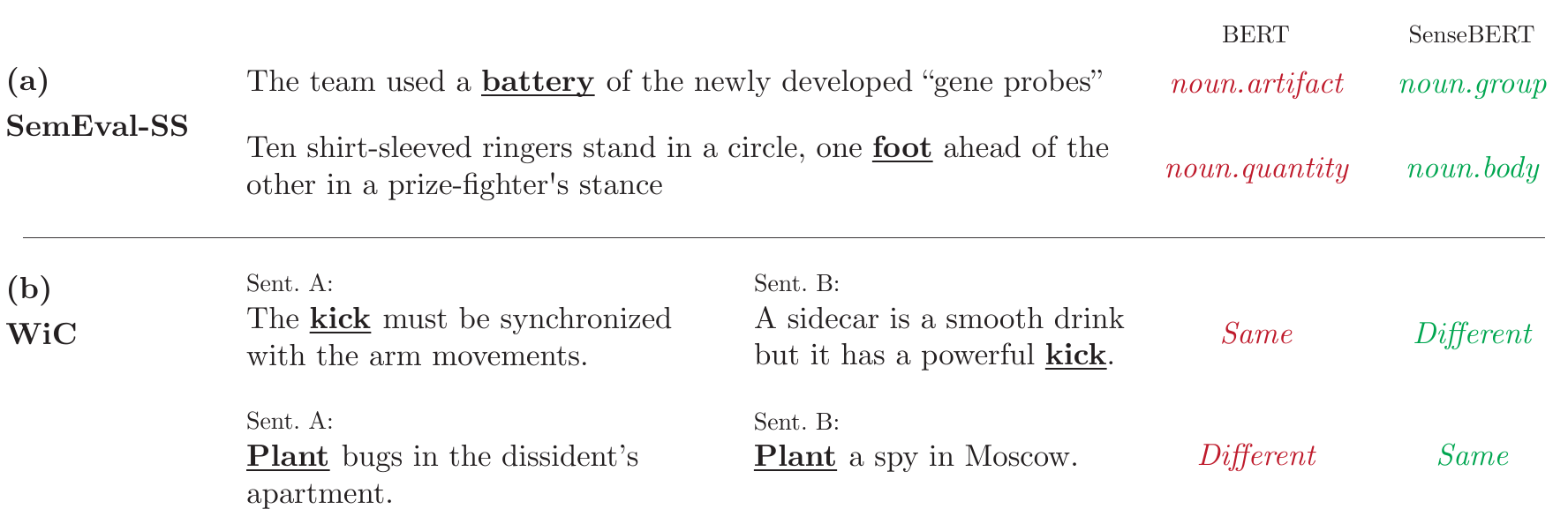}
		\vspace{-2mm} 
		\caption{Example entries of \textbf{(a)} the SemEval-SS task, where a model is to predict the supersense of the marked word, and \textbf{(b)} the Word in Context (WiC) task where a model must determine whether the underlined word is used in the same/different supersense within sentences A and B. In all displayed examples, taken from the corresponding development sets, \ourmodel~predicted the correct label while BERT failed to do so. A quantitative comparison between models is presented in table~\ref{tab:results}.}
		\label{fig:sota_ex}
	\end{figure*}
	
	We illustrate the resultant mapping in figure~\ref{fig:cluster}, showing a UMAP dimensionality reduction~\citep{mcinnes2018umap} of the rows of $S$, which corresponds to the different supersenses. A clear clustering according to the supersense part-of-speech is apparent in figure~\hyperref[fig:cluster]{~\ref{fig:cluster}(a)}. We further identify finer-grained semantic clusters, as shown for example in figure~\hyperref[fig:cluster]{~\ref{fig:cluster}(b)} and given in more detail in the supplementary materials.
	
	\ourmodel's semantic language model allows predicting a distribution over supersenses rather than over words in a masked position. Figure~\hyperref[fig:slm]{~\ref{fig:slm}(a)} shows the supersense probabilities assigned by \ourmodel~in several contexts, demonstrating the model's ability to assign semantically meaningful categories to the masked position.
	
	Finally, we demonstrate that SenseBERT enjoys an ability to view raw text at a lexical semantic level. Figure~\hyperref[fig:slm]{~\ref{fig:slm}(b)} shows example sentences and their supersense prediction by the pretrained model. Where a vanilla BERT would see only the words of the sentence ``Dan cooked a bass on the grill", \ourmodel~would also have access to the supersense abstraction: ``[Person] [created] [food] on the [artifact]". This sense-level perspective can help the model extract more knowledge from every training example, and to generalize semantically similar notions which do not share the same phrasing.

	\section{Lexical Semantics Experiments}
	\label{sec:results}
	
	In this section, we present quantitative evaluations of \ourmodel,~pre-trained as described in section~\ref{sec:method}. We test the model's performance  on a supersense-based variant of the SemEval WSD test sets standardized in~\citet{raganato2017word}, and on the Word in Context (WiC) task~\citep{pilehvar2018wic} (included in the recently introduced SuperGLUE benchmark~\citep{wang2019superglue}), both directly relying on the network's ability to perform lexical semantic categorization.


\subsection{Comparing Rare Words Supersense Prediction Methods}\label{sec:results:oov}
			
	We first report a comparison of the two methods described in section~\ref{sec:method:vocab} for predicting the supersenses of rare words which do not appear in BERT's original vocabulary. The first \emph{$60$K vocabulary} method enriches the vocabulary and the second \emph{average embedding} method predicts a supersense from the average embeddings of the sub-word tokens comprising an OOV word. 
	During fine-tuning, when encountering an OOV word we predict the supersenses from the rightmost sub-word token in the {$60$K vocabulary} method and from the average of the sub-word tokens in the {average embedding} method. 
	
	As shown in table~\ref{tab:oov}, both methods perform comparably on the SemEval supersense disambiguation task (see following subsection), yielding an improvement over the baseline of learning supersense information only for whole words in BERT's original $30$K-token vocabulary. We continue with the $60$K-token vocabulary for the rest of the experiments, but note the average embedding option as a viable competitor for predicting word-level semantics.

	\subsection{SemEval-SS: Supersense Disambiguation}\label{sec:results:sem}
		
	\begin{table*}[t]
		\begin{center}
			\begin{tabular}{lcccc}
				\toprule
				\bf  & \bf  SemEval-SS Frozen  & \bf SemEval-SS Fine-tuned & \bf Word in Context \\
				\midrule
				\bertbase & 65.1& 79.2 & -- \\
				\bertlarge  & 67.3 & 81.1 & 69.6\\
				\ourmodelbase& 75.6 & 83.0 & 70.3 \\
				\ourmodellarge  &  79.5 &  83.7 & 72.1 \\
				\bottomrule
			\end{tabular}
		\end{center}
		\caption{Results on a supersense variant of the SemEval WSD test set standardized in~\citet{raganato2017word}, which we denote SemEval-SS, and on the Word in Context (WiC) dataset~\citep{pilehvar2018wic} included in the recently introduced SuperGLUE benchmark~\citep{wang2019superglue}. 
			These tasks require a high level of lexical semantic understanding, as can be seen in the examples in figure~\ref{fig:sota_ex}. 
			For both tasks,  \ourmodel~demonstrates a clear improvement over BERT in the regular fine-tuning setup, where network weights are modified during training on the task.
			Notably, \ourmodellarge~achieves state of the art performance on the WiC task.
			In the SemEval-SS Frozen setting, we train a linear classifier over pretrained embeddings,  without changing the network weights. 
			The results show that \ourmodel~introduces a dramatic improvement in this setting, implying that its word-sense aware pre-training (section~\ref{sec:method}) yields embeddings that carries lexical semantic information which is  easily extractable for the benefits of downstream tasks. Results for BERT on the SemEval-SS task are attained by employing the published pre-trained BERT models, 
			and the \bertlarge~result on  WiC is taken from the baseline scores published on the SuperGLUE benchmark~\citep{wang2019superglue} (no result  has been published for \bertbase).}
		\label{tab:results}
	\end{table*} 
	\begin{table}
		\centering
		\begin{tabular}[t]{l c}
			\toprule
			\bf & \bf Word in Context \\
			\midrule
			ELMo$\dagger$ & 57.7 \\
			BERT sense embeddings $^{\dagger\dagger}$ & 67.7 \\
			\bertlarge$^\ddagger$ & 69.6 \\
			RoBERTa$^{\ddagger\ddagger}$ & 69.9\\
			KnowBERT-W+W$^\diamond$ & {70.9} \\
			SenseBERT & \textbf{72.1}\\
			\bottomrule
		\end{tabular}
		\caption{Test set results for the WiC dataset. \\
			$^\dagger$\citet{pilehvar2018wic} \\
			$^{\dagger\dagger}$\citet{loureiro2019language}\\
			$^{\ddagger}$\citet{wang2019superglue}\\
			$^{\ddagger\ddagger}$\citet{liu2019roberta}\\
			$^{\diamond}$\citet{peters2019knowledge}
		}
		\label{tab:wic}
	\end{table}
	We test \ourmodel~on a Word Supersense Disambiguation task, a coarse grained variant of the common WSD task. We use
	SemCor~\citep{miller1993semantic} as our training dataset ($226,036$ annotated examples), and the SenseEval~\citep{edmonds2001senseval,snyder2004english} / SemEval~\citep{pradhan2007semeval,navigli2013semeval,moro2015semeval} suite for evaluation (overall $7253$ annotated examples), following~\citet{raganato2017word}. For each word in both training and test sets, we change its fine-grained sense label to its corresponding WordNet supersense, and therefore train the network to predict a given word's supersense. We name this Supersense disambiguation task SemEval-SS. See figure~\hyperref[fig:sota_ex]{~\ref{fig:sota_ex}(a)} for an example from this modified data set.
	
	We show results on the SemEval-SS task for two different training schemes. In the first, we trained a linear classifier over the `frozen' output embeddings of the examined model -- we do not change the the trained \ourmodel's parameters in this scheme. This Frozen setting is a test for the amount of basic lexical semantics readily present in the pre-trained model, easily extricable by further downstream tasks (reminiscent of the semantic probes employed in~\citet{hewitt2019structural,coenen2019visualizing}.
	
	In the second training scheme we fine-tuned the examined model on the task, allowing its parameters to change during training (see full training details in the supplementary materials). 
	Results attained by employing this training method reflect the model's potential to acquire word-supersense information given its pre-training.	
	
	Table~\ref{tab:results} shows a comparison between vanilla BERT and \ourmodel~on the supersense disambiguation task. 
	Our semantic level pre-training signal clearly yields embeddings with enhanced word-meaning awareness, relative to embeddings trained with BERT's vanilla word-level signal. \ourmodelbase~improves the score of \bertbase~in the Frozen setting by over $10$ points and \ourmodellarge~ improves that of \bertlarge~ by over $12$ points, demonstrating competitive results even without fine-tuning.
	In the setting of model fine-tuning, we see a clear demonstration of the model's ability to learn word-level semantics, as \ourmodelbase~surpasses the score of \bertlarge~by $2$ points.
	
	\begin{table*}[t]\small
	\begin{center}
		\begin{tabular}{lccccccccc}
			\toprule
			\bf  & \bf  Score  & \bf CoLA & \bf SST-2 & \bf MRPC & \bf STS-B & \bf QQP & \bf MNLI & \bf QNLI & \bf RTE \\
			\midrule
			\bertbase\ \textsc{(Ours)} & 77.5 & 50.1 & 92.6 & 88.7/84.3 & 85.7/84.6 & 71.0/88.9 & 83.6 & 89.4 & 67.9 \\
			\ourmodelbase& 77.9 & 54.6 & 92.2 & 89.2/85.2 & 83.5/82.3 & 70.3/88.8 &	83.6 & 90.6 & 67.5 \\
			\bottomrule
		\end{tabular}
	\end{center}
	\caption{Results on the GLUE benchmark test set.}
	\label{tab:glue}
\end{table*}
	
	\subsection{Word in Context (WiC) Task}

	We test our model on the recently introduced WiC binary classification task. Each instance in WiC has a target word $w$ for which two contexts are provided, each invoking a specific meaning of $w$. The task is to determine whether the occurrences of $w$ in the two contexts share the same meaning or not, clearly requiring an ability to identify the word's semantic category. 
	The WiC task is defined over supersenses~\citep{pilehvar2018wic} -- the negative examples include a word used in two different supersenses and the positive ones include a word used in the same supersense. See figure~\hyperref[fig:sota_ex]{~\ref{fig:sota_ex}(b)} for an example from this data set.

	Results on the WiC task comparing \ourmodel~to vanilla BERT are shown in table~\ref{tab:results}. \ourmodelbase~surpasses a larger vanilla model, \bertlarge. As shown in table~\ref{tab:wic}, a single \ourmodellarge~model achieves the state of the art score in this task, demonstrating unprecedented lexical semantic awareness. 
	
	
	\subsection{GLUE}
	
	The General Language Understanding Evaluation (GLUE; \citet{wang2018glue}) benchmark is a popular testbed for language understanding models. It consists of 9 different NLP tasks, covering different linguistic phenomena. We evaluate our model on GLUE, in order to verify that \ourmodel~gains its lexical semantic knowledge without compromising performance on other downstream tasks. Due to slight differences in the data used for pretraining \textsc{BERT} and \ourmodel  \  (BookCorpus is not publicly available), we trained a \bertbase~model with the same data used for our models. \bertbase~and \ourmodelbase~were both finetuned using the exact same procedures and hyperparameters. The results are presented in table~\ref{tab:glue}. Indeed, \ourmodel~performs on par with BERT, achieving an overall score of 77.9, compared to 77.5 achieved by \bertbase.

	\section{Conclusion}
	\label{sec:conclusion}
	
	We introduce lexical semantic information into a neural language model's pre-training objective. 
	This results in a boosted word-level semantic awareness of the resultant model, named \ourmodel, which considerably outperforms a vanilla BERT on a SemEval based Supersense Disambiguation task and achieves state of the art results on the Word in Context task.
	This improvement was obtained without human annotation, but rather by harnessing an external linguistic knowledge source. Our work indicates that  semantic signals extending beyond the lexical level can be similarly introduced at the pre-training stage, allowing the network to elicit further insight without human supervision.
	
	\section*{Acknowledgments}
	We acknowledge useful comments and assistance from our colleagues at AI21 Labs. We would also like to thank the anonymous reviewers for their valuable feedback.
	%
	%

	%

    
	\bibliography{refs}

\begin{thebibliography}{39}
\expandafter\ifx\csname natexlab\endcsname\relax\def\natexlab#1{#1}\fi

\bibitem[{Arora et~al.(2018)Arora, Li, Liang, Ma, and
  Risteski}]{arora2018linear}
Sanjeev Arora, Yuanzhi Li, Yingyu Liang, Tengyu Ma, and Andrej Risteski. 2018.
\newblock \href {https://doi.org/10.1162/tacl_a_00034} {Linear algebraic
  structure of word senses, with applications to polysemy}.
\newblock \emph{Transactions of the Association for Computational Linguistics},
  6:483--495.

\bibitem[{Basile(2012)}]{basile2012super}
Pierpaolo Basile. 2012.
\newblock Super-sense tagging using support vector machines and distributional
  features.
\newblock In \emph{International Workshop on Evaluation of Natural Language and
  Speech Tool for Italian}, pages 176--185. Springer.

\bibitem[{Chan et~al.(2019)Chan, Kitaev, Guu, Stern, and
  Uszkoreit}]{chan2019kermit}
William Chan, Nikita Kitaev, Kelvin Guu, Mitchell Stern, and Jakob Uszkoreit.
  2019.
\newblock \href {https://arxiv.org/abs/1906.01604} {{KERMIT}: Generative
  insertion-based modeling for sequences}.
\newblock \emph{arXiv preprint arXiv:1906.01604}.

\bibitem[{Chen et~al.(2014)Chen, Liu, and Sun}]{chen2014unified}
Xinxiong Chen, Zhiyuan Liu, and Maosong Sun. 2014.
\newblock \href {https://doi.org/10.3115/v1/D14-1110} {A unified model for word
  sense representation and disambiguation}.
\newblock In \emph{Proceedings of the 2014 Conference on Empirical Methods in
  Natural Language Processing ({EMNLP})}, pages 1025--1035, Doha, Qatar.
  Association for Computational Linguistics.

\bibitem[{Ciaramita and Johnson(2003)}]{ciaramita2003supersense}
Massimiliano Ciaramita and Mark Johnson. 2003.
\newblock \href {https://www.aclweb.org/anthology/W03-1022} {Supersense tagging
  of unknown nouns in {W}ord{N}et}.
\newblock In \emph{Proceedings of the 2003 Conference on Empirical Methods in
  Natural Language Processing}, pages 168--175.

\bibitem[{Devlin et~al.(2019)Devlin, Chang, Lee, and
  Toutanova}]{devlin2018bert}
Jacob Devlin, Ming-Wei Chang, Kenton Lee, and Kristina Toutanova. 2019.
\newblock \href {https://doi.org/10.18653/v1/N19-1423} {{BERT}: Pre-training of
  deep bidirectional transformers for language understanding}.
\newblock In \emph{Proceedings of the 2019 Conference of the North {A}merican
  Chapter of the Association for Computational Linguistics: Human Language
  Technologies, Volume 1 (Long and Short Papers)}, pages 4171--4186,
  Minneapolis, Minnesota. Association for Computational Linguistics.

\bibitem[{Edmonds and Cotton(2001)}]{edmonds2001senseval}
Philip Edmonds and Scott Cotton. 2001.
\newblock \href {https://www.aclweb.org/anthology/S01-1001} {{SENSEVAL}-2:
  Overview}.
\newblock In \emph{Proceedings of {SENSEVAL}-2 Second International Workshop on
  Evaluating Word Sense Disambiguation Systems}, pages 1--5, Toulouse, France.
  Association for Computational Linguistics.

\bibitem[{Hewitt and Manning(2019)}]{hewitt2019structural}
John Hewitt and Christopher~D. Manning. 2019.
\newblock \href {https://doi.org/10.18653/v1/N19-1419} {{A} structural probe
  for finding syntax in word representations}.
\newblock In \emph{Proceedings of the 2019 Conference of the North {A}merican
  Chapter of the Association for Computational Linguistics: Human Language
  Technologies, Volume 1 (Long and Short Papers)}, pages 4129--4138,
  Minneapolis, Minnesota. Association for Computational Linguistics.

\bibitem[{Iacobacci et~al.(2016)Iacobacci, Pilehvar, and
  Navigli}]{iacobacci2016embeddings}
Ignacio Iacobacci, Mohammad~Taher Pilehvar, and Roberto Navigli. 2016.
\newblock \href {https://doi.org/10.18653/v1/P16-1085} {Embeddings for word
  sense disambiguation: An evaluation study}.
\newblock In \emph{Proceedings of the 54th Annual Meeting of the Association
  for Computational Linguistics (Volume 1: Long Papers)}, pages 897--907,
  Berlin, Germany. Association for Computational Linguistics.

\bibitem[{Inan et~al.(2017)Inan, Khosravi, and Socher}]{inan2016tying}
Hakan Inan, Khashayar Khosravi, and Richard Socher. 2017.
\newblock \href {https://openreview.net/pdf?id=r1aPbsFle} {Tying word vectors
  and word classifiers: A loss framework for language modeling}.
\newblock In \emph{ICLR}.

\bibitem[{Kilgarriff(1997)}]{kilgarriff1997don}
Adam Kilgarriff. 1997.
\newblock I don’t believe in word senses.
\newblock \emph{Computers and the Humanities}, 31(2):91--113.

\bibitem[{Le et~al.(2018)Le, Postma, Urbani, and Vossen}]{le2018deep}
Minh Le, Marten Postma, Jacopo Urbani, and Piek Vossen. 2018.
\newblock \href {https://www.aclweb.org/anthology/C18-1030} {A deep dive into
  word sense disambiguation with {LSTM}}.
\newblock In \emph{Proceedings of the 27th International Conference on
  Computational Linguistics}, pages 354--365, Santa Fe, New Mexico, USA.
  Association for Computational Linguistics.

\bibitem[{Liu et~al.(2019)Liu, Ott, Goyal, Du, Joshi, Chen, Levy, Lewis,
  Zettlemoyer, and Stoyanov}]{liu2019roberta}
Yinhan Liu, Myle Ott, Naman Goyal, Jingfei Du, Mandar Joshi, Danqi Chen, Omer
  Levy, Mike Lewis, Luke Zettlemoyer, and Veselin Stoyanov. 2019.
\newblock \href {https://arxiv.org/abs/1907.11692} {{RoBERTa}: A robustly
  optimized bert pretraining approach}.
\newblock \emph{arXiv preprint arXiv:1907.11692}.

\bibitem[{Loureiro and Jorge(2019)}]{loureiro2019language}
Daniel Loureiro and Al{\'\i}pio Jorge. 2019.
\newblock \href {https://doi.org/10.18653/v1/P19-1569} {Language modelling
  makes sense: Propagating representations through {W}ord{N}et for
  full-coverage word sense disambiguation}.
\newblock In \emph{Proceedings of the 57th Annual Meeting of the Association
  for Computational Linguistics}, pages 5682--5691, Florence, Italy.
  Association for Computational Linguistics.

\bibitem[{McInnes et~al.(2018)McInnes, Healy, and Melville}]{mcinnes2018umap}
Leland McInnes, John Healy, and James Melville. 2018.
\newblock \href {https://arxiv.org/abs/1802.03426} {{UMAP}: Uniform manifold
  approximation and projection for dimension reduction}.
\newblock \emph{arXiv preprint arXiv:1802.03426}.

\bibitem[{Mikolov et~al.(2013)Mikolov, Sutskever, Chen, Corrado, and
  Dean}]{mikolov2013distributed}
Tomas Mikolov, Ilya Sutskever, Kai Chen, Greg~S Corrado, and Jeff Dean. 2013.
\newblock \href
  {http://papers.nips.cc/paper/5021-distributed-representations-of-words-and-phrases-and-their-compositionality.pdf}
  {Distributed representations of words and phrases and their
  compositionality}.
\newblock In \emph{Advances in Neural Information Processing Systems 26}, pages
  3111--3119. Curran Associates, Inc.

\bibitem[{Miller(1998)}]{miller1998wordnet}
George~A Miller. 1998.
\newblock \emph{WordNet: An electronic lexical database}.
\newblock MIT press.

\bibitem[{Miller et~al.(1993)Miller, Leacock, Tengi, and
  Bunker}]{miller1993semantic}
George~A. Miller, Claudia Leacock, Randee Tengi, and Ross~T. Bunker. 1993.
\newblock \href {https://www.aclweb.org/anthology/H93-1061} {A semantic
  concordance}.
\newblock In \emph{{H}uman {L}anguage {T}echnology: Proceedings of a Workshop
  Held at Plainsboro, New Jersey, March 21-24, 1993}.

\bibitem[{Moro and Navigli(2015)}]{moro2015semeval}
Andrea Moro and Roberto Navigli. 2015.
\newblock \href {https://doi.org/10.18653/v1/S15-2049} {{S}em{E}val-2015 task
  13: Multilingual all-words sense disambiguation and entity linking}.
\newblock In \emph{Proceedings of the 9th International Workshop on Semantic
  Evaluation ({S}em{E}val 2015)}, pages 288--297, Denver, Colorado. Association
  for Computational Linguistics.

\bibitem[{Navigli(2009)}]{navigli2009word}
Roberto Navigli. 2009.
\newblock \href {https://doi.org/10.1145/1459352.1459355} {Word sense
  disambiguation: A survey}.
\newblock \emph{ACM Comput. Surv.}, 41(2).

\bibitem[{Navigli et~al.(2013)Navigli, Jurgens, and
  Vannella}]{navigli2013semeval}
Roberto Navigli, David Jurgens, and Daniele Vannella. 2013.
\newblock \href {https://www.aclweb.org/anthology/S13-2040} {{S}em{E}val-2013
  task 12: Multilingual word sense disambiguation}.
\newblock In \emph{Second Joint Conference on Lexical and Computational
  Semantics (*{SEM}), Volume 2: Proceedings of the Seventh International
  Workshop on Semantic Evaluation ({S}em{E}val 2013)}, pages 222--231, Atlanta,
  Georgia, USA. Association for Computational Linguistics.

\bibitem[{Pennington et~al.(2014)Pennington, Socher, and
  Manning}]{pennington2014glove}
Jeffrey Pennington, Richard Socher, and Christopher Manning. 2014.
\newblock \href {https://doi.org/10.3115/v1/D14-1162} {{G}love: Global vectors
  for word representation}.
\newblock In \emph{Proceedings of the 2014 Conference on Empirical Methods in
  Natural Language Processing ({EMNLP})}, pages 1532--1543, Doha, Qatar.
  Association for Computational Linguistics.

\bibitem[{Peters et~al.(2018)Peters, Neumann, Iyyer, Gardner, Clark, Lee, and
  Zettlemoyer}]{peters2018deep}
Matthew Peters, Mark Neumann, Mohit Iyyer, Matt Gardner, Christopher Clark,
  Kenton Lee, and Luke Zettlemoyer. 2018.
\newblock \href {https://doi.org/10.18653/v1/N18-1202} {Deep contextualized
  word representations}.
\newblock In \emph{Proceedings of the 2018 Conference of the North {A}merican
  Chapter of the Association for Computational Linguistics: Human Language
  Technologies, Volume 1 (Long Papers)}, pages 2227--2237, New Orleans,
  Louisiana. Association for Computational Linguistics.

\bibitem[{Peters et~al.(2019)Peters, Neumann, Logan, Schwartz, Joshi, Singh,
  and Smith}]{peters2019knowledge}
Matthew~E. Peters, Mark Neumann, Robert Logan, Roy Schwartz, Vidur Joshi,
  Sameer Singh, and Noah~A. Smith. 2019.
\newblock \href {https://doi.org/10.18653/v1/D19-1005} {Knowledge enhanced
  contextual word representations}.
\newblock In \emph{Proceedings of the 2019 Conference on Empirical Methods in
  Natural Language Processing and the 9th International Joint Conference on
  Natural Language Processing (EMNLP-IJCNLP)}, pages 43--54, Hong Kong, China.
  Association for Computational Linguistics.

\bibitem[{Pilehvar and Camacho-Collados(2019)}]{pilehvar2018wic}
Mohammad~Taher Pilehvar and Jose Camacho-Collados. 2019.
\newblock \href {https://doi.org/10.18653/v1/N19-1128} {{W}i{C}: the
  word-in-context dataset for evaluating context-sensitive meaning
  representations}.
\newblock In \emph{Proceedings of the 2019 Conference of the North {A}merican
  Chapter of the Association for Computational Linguistics: Human Language
  Technologies, Volume 1 (Long and Short Papers)}, pages 1267--1273,
  Minneapolis, Minnesota. Association for Computational Linguistics.

\bibitem[{Pradhan et~al.(2007)Pradhan, Loper, Dligach, and
  Palmer}]{pradhan2007semeval}
Sameer Pradhan, Edward Loper, Dmitriy Dligach, and Martha Palmer. 2007.
\newblock \href {https://www.aclweb.org/anthology/S07-1016} {{S}em{E}val-2007
  task-17: {E}nglish lexical sample, {SRL} and all words}.
\newblock In \emph{Proceedings of the Fourth International Workshop on Semantic
  Evaluations ({S}em{E}val-2007)}, pages 87--92, Prague, Czech Republic.
  Association for Computational Linguistics.

\bibitem[{Press and Wolf(2017)}]{press2016using}
Ofir Press and Lior Wolf. 2017.
\newblock \href {https://www.aclweb.org/anthology/E17-2025} {Using the output
  embedding to improve language models}.
\newblock In \emph{Proceedings of the 15th Conference of the {E}uropean Chapter
  of the Association for Computational Linguistics: Volume 2, Short Papers},
  pages 157--163, Valencia, Spain. Association for Computational Linguistics.

\bibitem[{Radford et~al.(2019)Radford, Wu, Child, Luan, Amodei, and
  Sutskever}]{radford2019language}
Alec Radford, Jeffrey Wu, Rewon Child, David Luan, Dario Amodei, and Ilya
  Sutskever. 2019.
\newblock Language models are unsupervised multitask learners.

\bibitem[{Raganato et~al.(2017)Raganato, Camacho-Collados, and
  Navigli}]{raganato2017word}
Alessandro Raganato, Jose Camacho-Collados, and Roberto Navigli. 2017.
\newblock \href {https://www.aclweb.org/anthology/E17-1010} {Word sense
  disambiguation: A unified evaluation framework and empirical comparison}.
\newblock In \emph{Proceedings of the 15th Conference of the {E}uropean Chapter
  of the Association for Computational Linguistics: Volume 1, Long Papers},
  pages 99--110, Valencia, Spain. Association for Computational Linguistics.

\bibitem[{Reif et~al.(2019)Reif, Yuan, Wattenberg, Viegas, Coenen, Pearce, and
  Kim}]{coenen2019visualizing}
Emily Reif, Ann Yuan, Martin Wattenberg, Fernanda~B Viegas, Andy Coenen, Adam
  Pearce, and Been Kim. 2019.
\newblock \href
  {http://papers.nips.cc/paper/9065-visualizing-and-measuring-the-geometry-of-bert.pdf}
  {Visualizing and measuring the geometry of {BERT}}.
\newblock In \emph{Advances in Neural Information Processing Systems 32}, pages
  8594--8603. Curran Associates, Inc.

\bibitem[{Rothe and Sch{\"u}tze(2015)}]{rothe2015autoextend}
Sascha Rothe and Hinrich Sch{\"u}tze. 2015.
\newblock \href {https://doi.org/10.3115/v1/P15-1173} {{A}uto{E}xtend:
  Extending word embeddings to embeddings for synsets and lexemes}.
\newblock In \emph{Proceedings of the 53rd Annual Meeting of the Association
  for Computational Linguistics and the 7th International Joint Conference on
  Natural Language Processing (Volume 1: Long Papers)}, pages 1793--1803,
  Beijing, China. Association for Computational Linguistics.

\bibitem[{Schneider(2014)}]{schneider2014lexical}
Nathan Schneider. 2014.
\newblock Lexical semantic analysis in natural language text.
\newblock \emph{Unpublished Doctoral Dissertation, Carnegie Mellon University}.

\bibitem[{Schneider and Smith(2015)}]{schneider2015corpus}
Nathan Schneider and Noah~A. Smith. 2015.
\newblock \href {https://doi.org/10.3115/v1/N15-1177} {A corpus and model
  integrating multiword expressions and supersenses}.
\newblock In \emph{Proceedings of the 2015 Conference of the North {A}merican
  Chapter of the Association for Computational Linguistics: Human Language
  Technologies}, pages 1537--1547, Denver, Colorado. Association for
  Computational Linguistics.

\bibitem[{Snyder and Palmer(2004)}]{snyder2004english}
Benjamin Snyder and Martha Palmer. 2004.
\newblock \href {https://www.aclweb.org/anthology/W04-0811} {The {E}nglish
  all-words task}.
\newblock In \emph{Proceedings of {SENSEVAL}-3, the Third International
  Workshop on the Evaluation of Systems for the Semantic Analysis of Text},
  pages 41--43, Barcelona, Spain. Association for Computational Linguistics.

\bibitem[{Vaswani et~al.(2017)Vaswani, Shazeer, Parmar, Uszkoreit, Jones,
  Gomez, Kaiser, and Polosukhin}]{vaswani2017attention}
Ashish Vaswani, Noam Shazeer, Niki Parmar, Jakob Uszkoreit, Llion Jones,
  Aidan~N Gomez, \L~ukasz Kaiser, and Illia Polosukhin. 2017.
\newblock \href
  {http://papers.nips.cc/paper/7181-attention-is-all-you-need.pdf} {Attention
  is all you need}.
\newblock In \emph{Advances in Neural Information Processing Systems 30}, pages
  5998--6008. Curran Associates, Inc.

\bibitem[{Wang et~al.(2019)Wang, Pruksachatkun, Nangia, Singh, Michael, Hill,
  Levy, and Bowman}]{wang2019superglue}
Alex Wang, Yada Pruksachatkun, Nikita Nangia, Amanpreet Singh, Julian Michael,
  Felix Hill, Omer Levy, and Samuel Bowman. 2019.
\newblock \href
  {http://papers.nips.cc/paper/8589-superglue-a-stickier-benchmark-for-general-purpose-language-understanding-systems.pdf}
  {{SuperGLUE}: A stickier benchmark for general-purpose language understanding
  systems}.
\newblock In \emph{Advances in Neural Information Processing Systems 32}, pages
  3266--3280. Curran Associates, Inc.

\bibitem[{Wang et~al.(2018)Wang, Singh, Michael, Hill, Levy, and
  Bowman}]{wang2018glue}
Alex Wang, Amanpreet Singh, Julian Michael, Felix Hill, Omer Levy, and Samuel
  Bowman. 2018.
\newblock \href {https://doi.org/10.18653/v1/W18-5446} {{GLUE}: A multi-task
  benchmark and analysis platform for natural language understanding}.
\newblock In \emph{Proceedings of the 2018 {EMNLP} Workshop {B}lackbox{NLP}:
  Analyzing and Interpreting Neural Networks for {NLP}}, pages 353--355,
  Brussels, Belgium. Association for Computational Linguistics.

\bibitem[{Yang et~al.(2019)Yang, Dai, Yang, Carbonell, Salakhutdinov, and
  Le}]{yang2019xlnet}
Zhilin Yang, Zihang Dai, Yiming Yang, Jaime Carbonell, Russ~R Salakhutdinov,
  and Quoc~V Le. 2019.
\newblock \href
  {http://papers.nips.cc/paper/8812-xlnet-generalized-autoregressive-pretraining-for-language-understanding.pdf}
  {{XLNet}: Generalized autoregressive pretraining for language understanding}.
\newblock In \emph{Advances in Neural Information Processing Systems 32}, pages
  5753--5763. Curran Associates, Inc.

\bibitem[{Yuan et~al.(2016)Yuan, Richardson, Doherty, Evans, and
  Altendorf}]{yuan2016semi}
Dayu Yuan, Julian Richardson, Ryan Doherty, Colin Evans, and Eric Altendorf.
  2016.
\newblock \href {https://www.aclweb.org/anthology/C16-1130} {Semi-supervised
  word sense disambiguation with neural models}.
\newblock In \emph{Proceedings of {COLING} 2016, the 26th International
  Conference on Computational Linguistics: Technical Papers}, pages 1374--1385,
  Osaka, Japan. The COLING 2016 Organizing Committee.

\end{thebibliography}
	\bibliographystyle{acl_natbib}
	
	\appendix 
	
	\begin{figure*}[h]
	\centering
	\includegraphics[width=\linewidth]{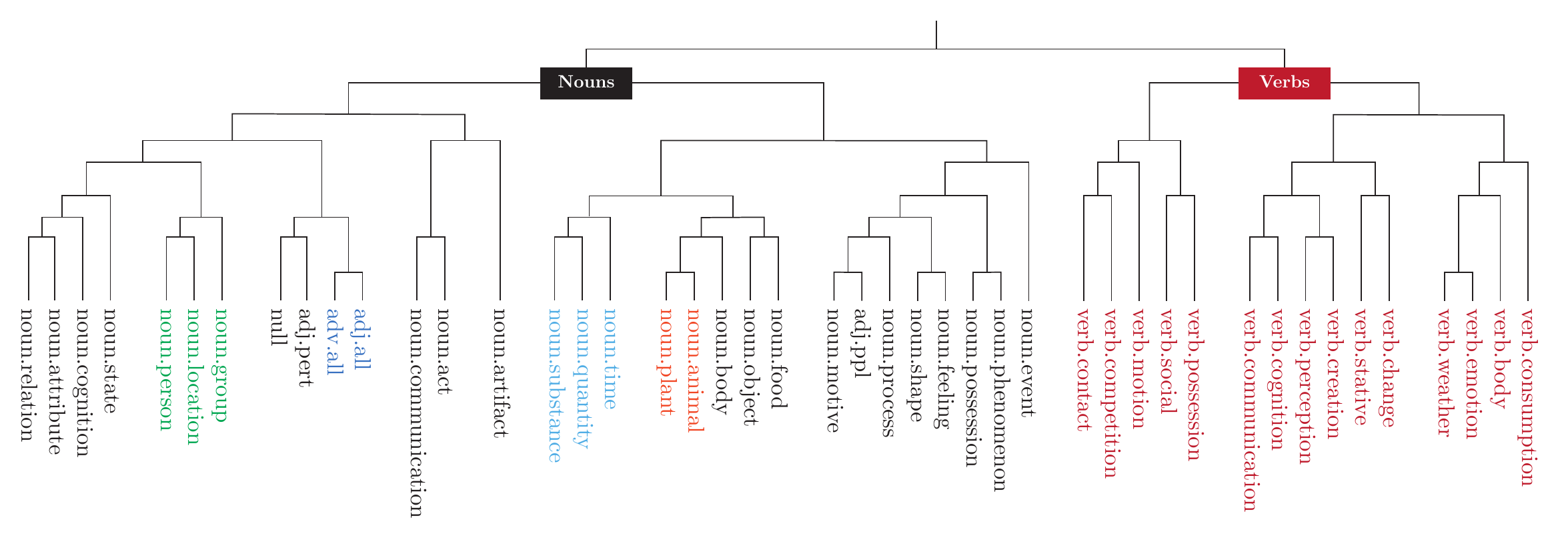}
	\vspace{-2mm} 
	\caption{Dendrogram visualization of an Agglomerative hierarchical clustering over the supersense vectors (rows of the classifier S) learned by SenseBERT.}
	\label{fig:dendogram}
\end{figure*}

\section{Supersenses and Their Representation in SenseBERT}\label{app:super}

We present in table~\ref{fig:ss} a comprehensive list of WordNet supersenses, as they appear in the WordNet documentation. 
In fig.~\ref{fig:dendogram} we present a Dendrogram of an Agglomerative hierarchical clustering over the supersense embedding vectors learned by SenseBERT in pre-training. The clustering shows a clear separation between Noun senses and Verb senses. Furthermore, we can observe that semantically related supersenses are clustered together (i.e, noun.animal and noun.plant).

\begin{table*}[t]
	\begin{center}
		{\small
			\hfill{}
			\resizebox{1.0\textwidth}{!}{
				\begin{tabular}{l|l|l|l}
					\textbf{Name} & \textbf{Content}& \textbf{Name} & \textbf{Content}\\
					\hline
					adj.all & All adjective clusters & noun.quantity  & Nouns denoting quantities and units \\
					&&& of measure\\ \cline{1-4}
					adj.pert & Relational adjectives (pertainyms) & noun.relation & Nouns denoting relations between  \\
					&&& people or things or ideas\\ \cline{1-4}
					adv.all & All adverbs & noun.shape & Nouns denoting two and three\\
					&&& dimensional shapes\\ \cline{1-4}
					noun.Tops & Unique beginner for nouns & noun.state & Nouns denoting stable states of affairs \\ 
					&&& \\ \cline{1-4}
					noun.act & Nouns denoting acts or actions & noun.substance & Nouns denoting substances \\ 
					&&& \\ \cline{1-4}
					noun.animal & Nouns denoting animals & noun.time & Nouns denoting time and temporal \\
					&&& relations\\ \cline{1-4}
					noun.artifact & Nouns denoting man-made objects & verb.body & Verbs of grooming, dressing \\
					&&&  and bodily care\\ \cline{1-4}
					noun.attribute & Nouns denoting attributes of people & verb.change & Verbs of size, temperature change, \\
					& and objects &&  intensifying, etc. \\ \cline{1-4}
					noun.body & Nouns denoting body parts & verb.cognition & Verbs of thinking, judging, analyzing,\\
					& &&  doubting \\ \cline{1-4} 
					noun.cognition & Nouns denoting cognitive & verb.communication & Verbs of telling, asking, ordering,  \\
					& processes and contents &&  singing \\ \cline{1-4} 
					noun.communication & Nouns denoting communicative & verb.competition & Verbs of fighting, athletic activities \\
					& processes and contents & & \\ \cline{1-4}
					noun.event & Nouns denoting natural events &  verb.consumption & Verbs of eating and drinking \\
					& & & \\ \cline{1-4}
					noun.feeling & Nouns denoting feelings & verb.contact & Verbs of touching, hitting, tying, \\
					& and emotions & & digging \\ \cline{1-4}
					noun.food & Nouns denoting foods and drinks & verb.creation & Verbs of sewing, baking, painting, \\
					& && performing \\ \cline{1-4}
					noun.group & Nouns denoting groupings of people & verb.emotion & Verbs of feeling \\
					& or objects &&\\ \cline{1-4}
					noun.location & Nouns denoting spatial position & verb.motion & Verbs of walking, flying, swimming \\
					& &&\\ \cline{1-4}
					noun.motive & Nouns denoting goals  & verb.perception & Verbs of seeing, hearing, feeling \\ 
					& &&\\ \cline{1-4}
					noun.object & Nouns denoting natural objects & verb.possession & Verbs of buying, selling, owning \\
					&(not man-made)&&\\ \cline{1-4}
					noun.person & Nouns denoting people & verb.social & Verbs of political and social  \\
					&& & activities and events \\ \cline{1-4}
					noun.phenomenon & Nouns denoting natural phenomena & verb.stative & Verbs of being, having, spatial relations \\
					& &&\\ \cline{1-4}
					noun.plant & Nouns denoting plants & verb.weather & Verbs of raining, snowing, thawing, \\
					& && thundering\\ \cline{1-4}
					noun.possession & Nouns denoting possession & adj.ppl & Participial adjectives \\
					&and transfer of possession &&\\ \cline{1-4}
					noun.process & Nouns denoting natural processes & & \\
					& &&\\
		\end{tabular}}}
		\hfill{}
		\caption{ 
			A list of supersense categories from WordNet lexicographer. 
		}
		\label{fig:ss}
	\end{center}
\end{table*}

\section{Training Details}
\label{app:training}
As hyperparameters for the fine-tuning, we used $max\_seq\_length = 128$, chose learning
rates from $\{5e-6, 1e-5, 2e-5, 3e-5, 5e-5\}$,
batch sizes from $\{16, 32\}$, and fine-tuned up to $10$
epochs for all the datasets.

\end{document}